\documentclass[11pt, letterpaper, logo, onecolumn, copyright, numbering]{minimax}
\usepackage{etoolbox}

\usepackage[authoryear, sort&compress, round]{natbib}

\usepackage[inkscapeformat=png]{svg}

\usepackage[most, breakable, skins]{tcolorbox}
\usepackage{academicons}

\tcbuselibrary{skins}
\usepackage{lipsum}
\usepackage{tabularx}
\usepackage{afterpage}
\usepackage{booktabs}
\usepackage{subcaption}
\usepackage{makecell}
\usepackage{multirow}
\usepackage{multicol} 
\usepackage{array}
\usepackage{float}
\usepackage{listings, listings-rust}
\usepackage{fontawesome5}
\usepackage{amssymb,graphicx}
\usepackage[dvipsnames]{xcolor}
\usepackage{hyperref}
\usepackage{cleveref}
\usepackage{longtable}
\usepackage{graphicx}
\usepackage{pdflscape}
\usepackage{adjustbox}
\usepackage{tikz}
\usetikzlibrary{calc,positioning,chains,shapes,arrows,fit,decorations.pathmorphing,patterns,fadings,shadows,patterns.meta,arrows.meta}
\usepackage{wrapfig}
\usepackage{dialogue}
\usepackage{algorithm}
\usepackage{algorithmic}
\usepackage{colortbl}
\usepackage{mdframed}

\usepackage{listings}

\usepackage{CJKutf8}
\usepackage{tcolorbox}

\usepackage[dvipsnames]{xcolor}
\usepackage{multicol}

\usepackage{caption}
\usepackage{amsmath,amsfonts,bm}

\def\eqref#1{equation~\ref{#1}}

\def\1{\bm{1}}

\DeclareMathAlphabet{\mathsfit}{\encodingdefault}{\sfdefault}{m}{sl}
\SetMathAlphabet{\mathsfit}{bold}{\encodingdefault}{\sfdefault}{bx}{n}

\definecolor{ababcol}{HTML}{F14738}
\definecolor{myhailuo2}{HTML}{F97669}
\definecolor{querycol}{HTML}{7964E8}
\definecolor{goldanswercol}{HTML}{FFB43B}
\definecolor{otherscol}{HTML}{FC5BCF}
\definecolor{myhailuo3light}{HTML}{FFA9FA}
\definecolor{myhailuo1dark}{HTML}{FC8900}
\definecolor{myhailuo2dark}{HTML}{F14738}
\definecolor{myhailuo3dark}{HTML}{D12AAA}
\definecolor{myhailuo4dark}{HTML}{4C4DC2}

\definecolor{myhailuo1}{HTML}{FFB43B}
\definecolor{myhailuo2}{HTML}{F97669}
\definecolor{myhailuo3}{HTML}{FC5BCF}
\definecolor{myhailuo4}{HTML}{7964E8}

\definecolor{myhailuo1light}{HTML}{FFD085}
\definecolor{myhailuo2light}{HTML}{FFA19F}
\definecolor{myhailuo3light}{HTML}{FFA9FA}
\definecolor{myhailuo4light}{HTML}{BDACFB}

\colorlet{myorange}{Orange!20}
\colorlet{mygreen}{LimeGreen!25}
\colorlet{myyellow}{Yellow!30}
\colorlet{myblue}{CornflowerBlue!25}
\colorlet{mybrown}{RawSienna!25}
\colorlet{mypurple}{Orchid!25}
\colorlet{myred}{Red!60}
\colorlet{myorangefull}{YellowOrange!60}
\colorlet{mybrownfull}{RawSienna!60}

\colorlet{myorangethick}{Orange!40}
\colorlet{mygreenthick}{LimeGreen!50}
\colorlet{myyellowthick}{Yellow!60}
\colorlet{mybluethick}{CornflowerBlue!50}

\tcbset{
    showcase/.style={
        fonttitle=\large,
        colback=white!20,  
        colframe=black,   
        coltitle=white,   
        boxrule=0.5mm,    
        arc=2mm,          
        outer arc=2mm,    
        left=1mm,         
        right=1mm,        
        top=1mm,          
        bottom=1mm,       
        width=\textwidth, 
        before skip=0.1pt,
        after skip=0.1pt,
    },
    context/.style={
        fontupper=\scriptsize,
        fonttitle=\large,
        colframe=querycol,     
        coltitle=white,   
        colback=white,    
        boxrule=0.3mm,    
        arc=2mm,          
        outer arc=2mm,    
        left=1mm,         
        right=1mm,        
        top=1mm,          
        bottom=1mm,       
        before skip=1pt,
        after skip=0.1pt, 
    },
    query/.style={
        fontupper=\scriptsize,
        fontlower=\scriptsize,
        colframe=querycol,     
        coltitle=white,   
        colback=white,    
        boxrule=0.1mm,    
        arc=2mm,          
        outer arc=2mm,    
        left=1mm,         
        right=1mm,        
        top=1mm,          
        bottom=1mm,       
        before skip=1pt,
        after skip=0.1pt,
    },
    abab/.style={
        fontupper=\scriptsize,
        fonttitle=,
        colframe=ababcol, 
        coltitle=white,   
        boxrule=0.5mm,    
        arc=2mm,          
        outer arc=2mm,    
        left=1mm,         
        right=1mm,        
        top=1mm,          
        bottom=1mm,       
        width=0.33\textwidth, 
        before skip=0.1pt,
        after skip=0.1pt, 
    },
    others/.style={
        fontupper=\scriptsize,
        colframe=myhailuo3light, 
        coltitle=white,
        boxrule=0.5mm,    
        arc=2mm,          
        outer arc=2mm,    
        left=1mm,         
        right=1mm,        
        top=1mm,          
        bottom=1mm,       
        width=0.33\textwidth, 
        before skip=0.1pt,
        after skip=0.1pt, 
    },
    goldanswer/.style={
        fontupper=\scriptsize,
        colframe=goldanswercol,     
        coltitle=white,   
        boxrule=0.5mm,    
        arc=2mm,          
        outer arc=2mm,    
        left=1mm,         
        right=1mm,        
        top=1mm,          
        bottom=1mm,       
        width=0.33\textwidth, 
        before skip=0.1pt,
        after skip=0.1pt, 
    },
}

\theoremstyle{plain}

\theoremstyle{definition}

\theoremstyle{remark}

\usepackage{CJKutf8}

\definecolor{medgray55}{gray}{0.55}
\definecolor{medgray}{gray}{0.7}
\definecolor{litegray}{gray}{0.9}
\definecolor{gblue}{RGB}{210, 227, 252}
\definecolor{gred}{RGB}{250, 210, 207}
\definecolor{gyellow}{RGB}{254, 239, 195}
\definecolor{ggreen}{RGB}{206, 234, 214}
\definecolor{gorange}{RGB}{254, 223, 200}

\definecolor{gblue9}{RGB}{23, 78, 166}
\definecolor{gred9}{RGB}{165, 14, 14}
\definecolor{gyellow9}{RGB}{227, 116, 0}
\definecolor{ggreen9}{RGB}{13, 101, 45}
\definecolor{gorange9}{RGB}{176, 96, 0}

\definecolor{myblue}{rgb}{0,0,1}
\definecolor{myred}{rgb}{1,0,0}
\definecolor{mylightgray}{gray}{0.95}

\definecolor{highlightblue}{HTML}{185ABC}

\makeatletter

\renewcommand\paragraph{\@startsection{paragraph}{4}{\z@}%
            {-2.5ex\@plus -1ex \@minus -.25ex}%
            {1.25ex \@plus .25ex}%
            {\itshape\normalsize\bfseries}}
\makeatother
\newcolumntype{L}[1]{>{\raggedright\let\newline\\\arraybackslash\hspace{0pt}}m{#1}}
\newcolumntype{C}[1]{>{\centering}m{#1}}

\newcolumntype{R}[1]{>{\raggedleft\let\newline\\\arraybackslash\hspace{0pt}}m{#1}}

\definecolor{ao}{rgb}{0.0, 0.0, 1.0}

\newcommand\vcent[1]{\vcenter{\hbox{#1}}}

\newcommand\loudspeaker[1][3]{\ensuremath{\vcent{\rule{.6ex}{.6ex}}\kern-.5ex%
  \vcent{\scalebox{.6}[1]{\rotatebox[origin=center]{90}{$\blacktriangle$}}}%
  \ifnum#1>0\relax\kern.05ex\vcent{\scalebox{.4}{\ttfamily)}}%
  \ifnum#1>1\relax\kern-.4ex\vcent{\scalebox{.56}{\ttfamily)}}%
  \ifnum#1>2\relax\kern-.55ex\vcent{\scalebox{.7}{\ttfamily)}}%
  \fi\fi\fi}%
}

\definecolor{green}{rgb}{0.9,0.9,0.9}

\newcommand{\method}{CISPO}

\makeatletter
\renewcommand\subparagraph{%
 \@startsection {subparagraph}{5}{\z@ }{3.25ex \@plus 1ex
 \@minus .2ex}{-1em}{\normalfont \normalsize \bfseries }}%
\makeatother

\let\cite\citep

\title{MiniMax-M1: Scaling Test-Time Compute Efficiently with Lightning Attention}

\reportnumber{}

\renewcommand{\today}{}

\author[*,1]{MiniMax\footnote{Please send correspondence to model@minimax.io.}}

\begin{abstract}
We introduce MiniMax-M1, the world's first open-weight, large-scale hybrid-attention reasoning model. MiniMax-M1 is powered by a hybrid Mixture-of-Experts (MoE) architecture combined with a lightning attention mechanism. The model is developed based on our previous MiniMax-Text-01 model~\citep{minimax2025minimax01}, which contains a total of 456 billion parameters with 45.9 billion parameters activated per token. The M1 model natively supports a context length of 1 million tokens, 8x the context size of DeepSeek R1. Furthermore, the lightning attention mechanism in MiniMax-M1 enables efficient scaling of test-time compute -- For example, compared to DeepSeek R1, M1 consumes 25\% of the FLOPs at a generation length of 100K tokens. These properties make M1 particularly suitable for complex tasks that require processing long inputs and thinking extensively.
MiniMax-M1 is trained using large-scale reinforcement learning (RL) on diverse problems ranging from traditional mathematical reasoning to sandbox-based, real-world software engineering environments. 
In addition to the inherent efficiency advantage of lightning attention for RL training, we propose \method{}, a novel RL algorithm to further enhance RL efficiency. \method{} clips importance sampling weights rather than token updates, outperforming other competitive RL variants.
Combining hybrid-attention and \method{} enables MiniMax-M1's full RL training on 512 H800 GPUs to complete in only three weeks, with a rental cost of just \$534,700.
We release two versions of MiniMax-M1 models with 40K and 80K thinking budgets respectively, where the 40K model represents an intermediate phase of the 80K training.
Experiments on standard benchmarks show that our models are comparable or superior to strong open-weight models such as the original DeepSeek-R1 and Qwen3-235B, with particular strengths in complex software engineering, tool utilization, and long-context tasks. 
Through efficient scaling of test-time compute, MiniMax-M1 serves as a strong foundation for next-generation language model agents to reason and tackle real-world challenges. We publicly release MiniMax-M1 at \href{https://github.com/MiniMax-AI/MiniMax-M1}{https://github.com/MiniMax-AI/MiniMax-M1}. 

\end{abstract}

\begin{document}

\maketitle

\label{sec:intro}
\begin{figure*} [h]
\begin{subfigure}[t]{0.7\textwidth}  
    \centering
    \includegraphics[width=\textwidth]{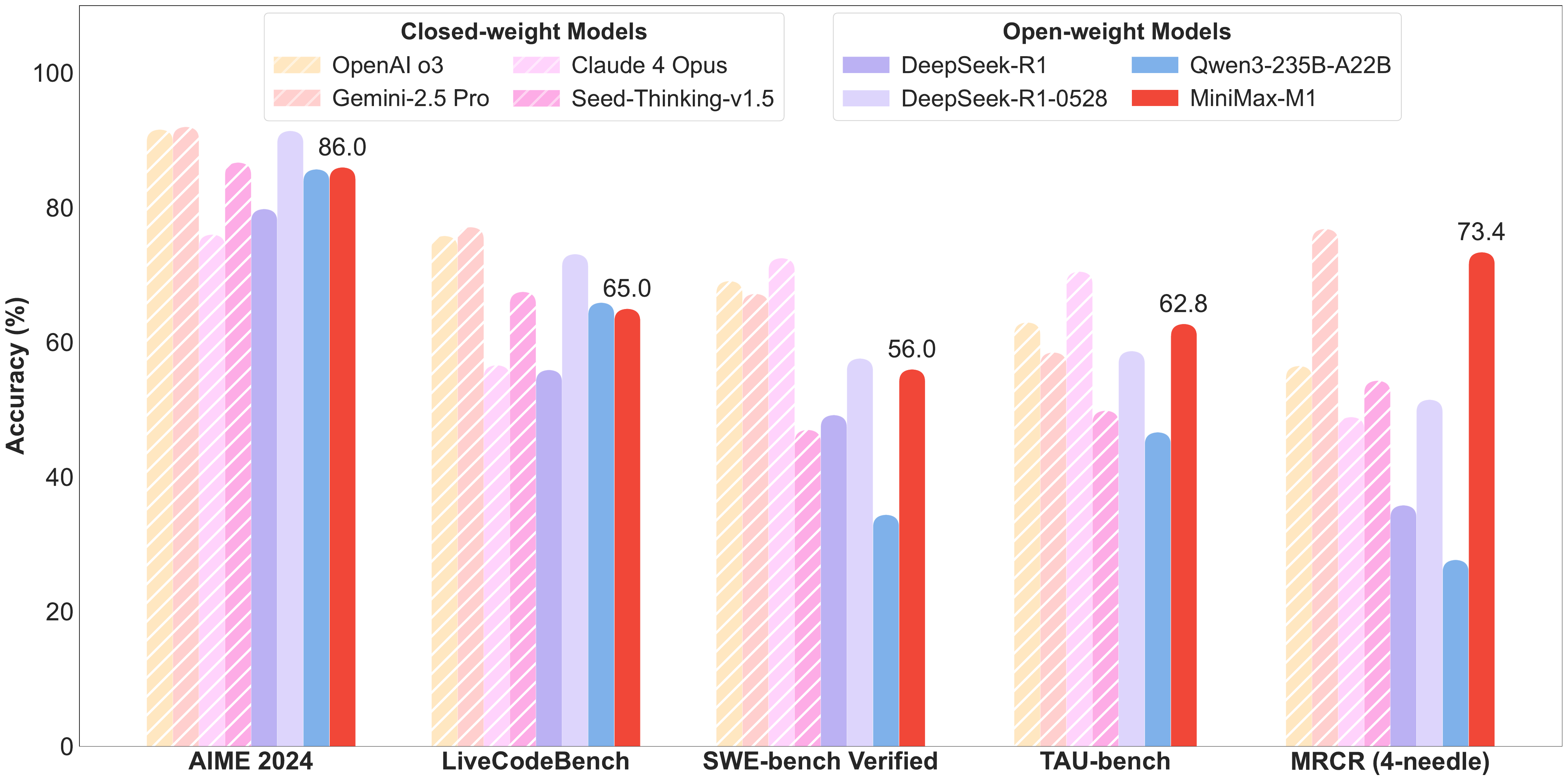}
\end{subfigure}
\hfill
\begin{subfigure}[t]{0.29\textwidth}  
    \centering
    \includegraphics[width=\textwidth]{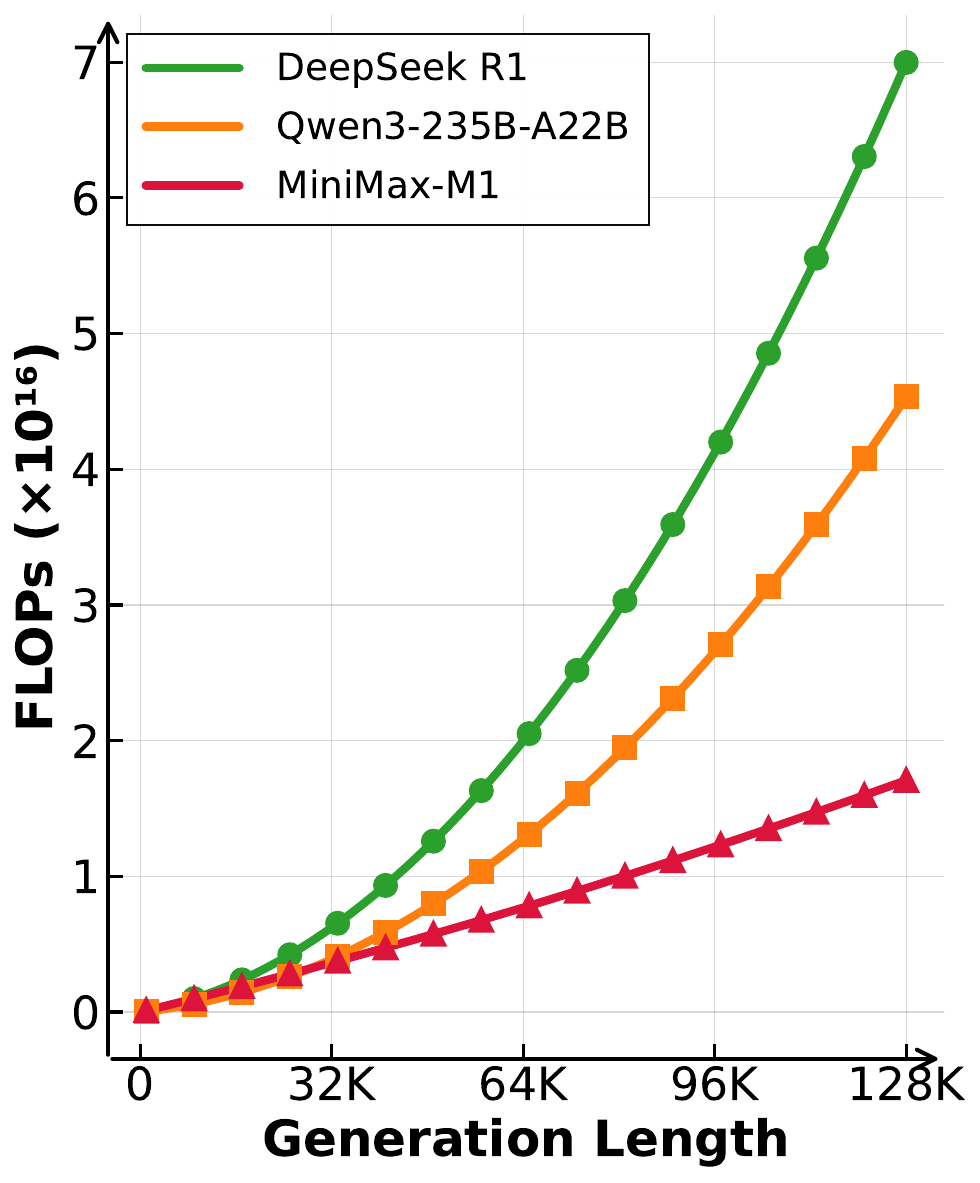}
\end{subfigure}
    \caption{\textbf{Left}: Benchmark performance comparison of leading commercial and open-weight models across competition-level mathematics, coding, software engineering, agentic tool use, and long-context understanding tasks. We use the MiniMax-M1-80k model here for MiniMax-M1. 
    \textbf{Right}: Theoretical inference FLOPs scaling with generation length (\# tokens).}

    \label{fig:perf-bars-flops}
\end{figure*}

\section{Introduction}

Large reasoning models (LRMs), such as OpenAI o1~\citep{openai2024o1} and DeepSeek-R1~\citep{deepseekai2025deepseekr1incentivizingreasoningcapability}, have demonstrated remarkable success by extending the length of reasoning through large-scale reinforcement learning (RL). In recent months, both the open-source community and commercial organizations have followed this trend, achieving significant advances on complex tasks such as Olympiad mathematics competitions and competitive programming~\citep{team2025kimi,anthropic2025claude37,deepmind2025geminipro,seed2025seed1,zeng2025simplerlzooinvestigatingtamingzero,yu2025dapoopensourcellmreinforcement,hu2025openreasonerzeroopensourceapproach}.
The success of LRMs has been primarily attributed to a new scaling dimension of test-time compute—As more FLOPs are dedicated to extended reasoning processes during generation, model performance shows consistent improvement, particularly for complex real-world applications~\citep{openai2025research,jimenez2024swebenchlanguagemodelsresolve}. 

However, continuously extending the reasoning process is challenging within the traditional transformer architecture~\citep{vaswani2017attention}, due to the inherent quadratic computational complexity of the softmax attention mechanism. While previous works have proposed various techniques to mitigate this issue—such as sparse attention~\citep{beltagy2020longformer,zaheer2020big,lu2025moba,yuan2025native}, linear attention~\citep{katharopoulos2020transformers,qin_cosformer_iclr_2021,choromanski2021rethinking,peng2021random,sun2023retentive,qin2022devil,zhen2022cosformer,qin2024you,qin2024various,peng_rwkv4_2024,sun2025linear,shen2024scaling,arora2024simple,zhang2024gated,du2025mom,he2024rodimus}, linear attention with delta decay~\cite{yang2024parallelizing,yang2024gated,peng2025rwkv}, state space models~\citep{gu2020hippo,gu2023how,s4, mamba,mamba2,glorioso2024zamba,ren2024samba,team2024jamba,gupta2022DSS}, and linear RNNs~\citep{hochreiter1997long,iclr18,gru,qin2023hierarchically,peng_rwkv_emnlp_2023,peng2024eagle,qin2024hgrn2,chou2024metala,siems2025deltaproduct,sun2024learning,von2025mesanet,behrouz2024titans}—these approaches have not been fully validated in large-scale reasoning models, and nearly all competitive LRMs to date still rely on traditional attention designs. An exception is the Hunyuan-T1 model~\citep{hunyuan_t1} that employs the Mamba architecture~\citep{mamba,mamba2}. However, this model is not open-sourced and few details are disclosed. 
In this work, we aim to build and open-source a large reasoning model that can efficiently scale up test-time compute and compete with the state-of-the-art reasoning models. 

We introduce MiniMax-M1, a reasoning model with a hybrid Mixture-of-Experts (MoE) architecture and Lightning Attention~\citep{qin2024lightning}, an I/O-aware implementation of a linear attention variant~\citep{qin2022devil}. MiniMax-M1 is developed based on our previous MiniMax-Text-01~\citep{minimax2025minimax01} model, and comprises 456 billion parameters in total, with 45.9 billion activations and 32 experts. In our attention design, a transformer block with softmax attention follows every seven transnormer blocks~\citep{qin2022devil} with lightning attention. This design theoretically enables efficient scaling of reasoning lengths to hundreds of thousands of tokens, as illustrated in Figure~\ref{fig:perf-bars-flops} (Right). For example, compared to DeepSeek R1, M1 consumes less than 50\% of the FLOPs at a generation length of 64K tokens, and approximately 25\% of the FLOPs at a length of 100K tokens. This substantial reduction in computational cost makes M1 significantly more efficient during both inference and large-scale RL training. Furthermore, owing to its lightning attention mechanism and in line with MiniMax-Text-01, our M1 model natively supports a context length of up to 1 million tokens -- eight times the context size of DeepSeek R1 and an order of magnitude greater than all open-weight LRMs available to date. These features make M1 particularly well-suited for addressing complex, real-world tasks that require processing long inputs and generating extended thinking. A comparison of the maximum input and output lengths of M1 and other leading models is demonstrated in Table~\ref{tab:context}.

\begin{table}[!t]
\caption{The maximum supported input length and output length (\# tokens) of different reasoning models. For Claude-4 we refer to the Claude-4-Opus model. ``DS-R1'' represents the latest \texttt{DeepSeek-R1-0528} model.}
\label{tab:context}
\centering
\begin{tabular}{lcccccc}
\toprule
                  & o3 & Gemini 2.5 Pro & Claude 4 & DS-R1 & Qwen3-235B & MiniMax-M1-80k \\
\midrule
Max Input  & 200K      & 1M             & 200K            & 128K        & 128K       & 1M             \\
Max Output & 100K      & 64K            & 32K             & 64K         & 32K        & 80K            \\
\bottomrule
\end{tabular}

\end{table}

To develop our M1 model, we first continue pretraining MiniMax-Text-01 on 7.5T tokens from a carefully curated, reasoning-intensive corpus. Subsequently, we perform supervised fine-tuning (SFT) to inject certain chain-of-thought (CoT)~\citep{wei2022chain} patterns, establishing a strong foundation for reinforcement learning, the core stage of M1 development.
Notably, our RL scaling with M1 is made efficient through innovations from two key perspectives: (1) We propose a novel RL algorithm, \method{}, which abandons the trust region constraint and instead clips the importance sampling weights to stabilize training. This approach always leverages all tokens for gradient computations, achieving enhanced efficiency compared to GRPO~\citep{shao2024deepseekmath} and DAPO~\citep{yu2025dapoopensourcellmreinforcement} empirically -- For example, on a controlled study based on Qwen2.5-32B models~\citep{qwen2025qwen25technicalreport}, \method{} achieves a 2x speedup compared to DAPO; (2) Although the hybrid-attention design in M1 naturally allows for efficient RL scaling, unique challenges arise when scaling RL with this architecture. For instance, we find a precision mismatch between the training and inference kernels of our architecture, which prevents reward growth during RL training. We develop targeted solutions to address these challenges and successfully scale up RL with this hybrid architecture.
In the end, our efficient RL framework enables us to complete a full RL run of MiniMax-M1 within 3 weeks using 512 H800 GPUs---equivalent to a rental cost of approximately \$0.53M USD.

In addition to methodological innovations, we curate a diverse set of problems and environments for RL training. Our data encompasses both verifiable and non-verifiable problems. For verifiable problems that are typically considered critical for reasoning learning, we not only include mathematical reasoning and competitive programming problems as commonly used in related works, but also leverage our previous data synthesis framework SynLogic~\citep{liu2025synlogic} to generate diverse logical reasoning problems spanning 41 distinct tasks. Furthermore, we construct sandboxes for complex software engineering (SE) environments derived from SWE-bench~\citep{jimenez2024swebenchlanguagemodelsresolve}, and conduct RL on real-world SE problems with execution-based rewards to improve M1's performance in challenging SE scenarios. Our unverifiable problems span a broad range of domains such as question answering and creative writing, where we use generative reward models to provide the feedback. 

We train two versions of MiniMax-M1 models with 40K and 80K tokens of maximum generation length respectively, which leads to two models MiniMax-M1-40k and MiniMax-M1-80k.
MiniMax-M1-80k outperforms MiniMax-M1-40k on complex mathematical and coding tasks, further demonstrating the benefits of scaling test-time compute. As shown in Figure~\ref{fig:perf-bars-flops} (Left), MiniMax-M1 surpasses previous leading open-weight models such as the original DeepSeek-R1 and Qwen-235B overall, with particular advantages in complex software engineering, tool-using, and long-context tasks.
Compared to the latest DeepSeek-R1-0528 model, MiniMax-M1 lags in mathematical and coding competitions but achieves comparable or superior performance in more realistic tool-using and long-context scenarios.
Notably, MiniMax-M1 outperforms Gemini 2.5 Pro on the agentic tool use benchmark TAU-Bench~\citep{yaotau}, and surpasses OpenAI o3 and Claude 4 Opus on long-context understanding benchmarks.
With efficient test-time scaling, we contend that MiniMax-M1 establishes a strong foundation for next-generation language model agents to address real-world challenges.

To facilitate collaboration and advancement in the field, we have made our models publicly available at GitHub and Hugging Face. They are now supported by both the \texttt{vLLM} and \texttt{Transformers} frameworks, with detailed deployment guides available at \href{https://github.com/MiniMax-AI/MiniMax-M1/blob/main/docs/vllm_deployment_guide.md}{vLLM} and \href{https://github.com/MiniMax-AI/MiniMax-M1/blob/main/docs/transformers_deployment_guide.md}{Transformers} respectively. This enables easy integration of MiniMax-M1 into modern inference pipelines. We also provide commercial standard API at \href{https://minimax.io}{minimax.io}.

\section{Preparation for Scalable RL: Continual Pretraining and SFT}

In this work, we focus on scaling up reinforcement learning to enhance reasoning capabilities of Minimax-Text-01. To facilitate scalable RL training, we first carry out continual pretraining of our base model to strengthen its intrinsic reasoning abilities. Subsequently, we perform a cold-start supervised fine-tuning (SFT) stage to inject specific reasoning patterns to the model, thereby providing a stronger foundation for the subsequent RL phase.

\subsection{Continual Pre-Training: Foundation for RL Scaling}

To enhance the reasoning and long context capabilities of the foundation model while ensuring diversity, we continue training the MiniMax-Text-01 model with additional 7.5T tokens with optimized data quality and mixture.

\noindent\textbf{Training Data.}
We refine our pretraining Web and PDF parsing mechanisms and enhance our heuristic cleaning rules to ensure a high recall rate for mathematical and code-related data. We prioritize the extraction of natural Question-Answer (QA) pairs from a diverse range of sources, including webpages, forums, and textbooks, while strictly avoiding the use of synthetic data. Additionally, we conduct semantic deduplication on the QA data to maintain its diversity and uniqueness. Furthermore, we increase the proportion of STEM (Science, Technology, Engineering, and Mathematics), code, book, and reasoning-related data to 70\%. This significantly enhances the foundation model's ability to handle complex tasks without compromising its other general capabilities.

\noindent\textbf{Training Recipe.}
We decrease the coefficient of the MoE auxiliary loss and adjust the parallel training strategy to support a larger training micro batch size, which mitigates the detrimental effects of the auxiliary loss on overall model performance. Based on MiniMax-Text-01, we continue training with a constant learning rate of 8e-5 for 2.5T tokens, followed by a decay schedule over 5T tokens down to 8e-6.

\noindent\textbf{Long Context Extension.}
For a hybrid-lightning architecture model with higher convergence complexity, we have observed that excessively aggressive extensions of the training length can lead to a sudden gradient explosion that may occur during the training process. This makes the optimization process extremely challenging. We attribute this to the parameter optimization of the earlier layers not keeping up with the changes in the later layers -- For lightning attention, the earlier and later layers have different decay rates, which makes the earlier layers focus more on local information. We alleviate this issue by adapting a smoother extension of context length across four stages, starting from a 32K context window length and ultimately extending the training context to 1M tokens.

\subsection{Supervised Fine-Tuning: Focused Alignment for Efficient RL}

After continual pretraining, we conduct Supervised Fine-Tuning (SFT) to instill desired behaviors like reflection-based Chain-of-Thought (CoT) reasoning using high-quality examples, creating a strong starting point for more efficient and stable RL in the next stage. Specifically, we curate data samples with long CoT responses. These data samples cover diverse domains such as math, coding, STEM, writing, QA, and multi-turn chat.  Math and coding samples account for around 60\% of all the data.

\section{Efficient RL Scaling: Algorithms and Lightning Attention} 
As shown in Figure~\ref{fig:perf-bars-flops} (Right), the M1 architecture demonstrates a clear efficiency advantage during inference. This naturally facilitates efficient RL scaling where increasingly longer responses are generated. However, as pioneers in scaling up RL with this hybrid architecture, we encounter unique challenges during the process, and the RL procedure can become unstable or even fail due to various issues.
To address these difficulties, we develop targeted solutions that enable us to successfully scale up RL training for M1. In addition, we propose a new RL algorithm that achieves greater RL efficiency compared to existing methods.
These dual contributions yield an efficient and scalable RL framework for training M1, where the complete training cycle requires 3 weeks on 512 H800 GPUs—equivalent to a rental cost of approximately \$0.53M USD.
In this section, we first provide general context on RL and present our novel RL algorithm, and then describe the specific challenges we face with the hybrid architecture, along with the solutions we devise to overcome them.

\subsection{Efficient RL Scaling with \method{}}
\label{sec:method}

\noindent\textbf{Background.}
For questions $q$ from a dataset $\mathcal{D}$, we denote $\pi$ as the policy model parameterized by $\theta$, and $o$ as the response generated by the policy.
PPO~\citep{schulman2017proximal} adopts the following objective to optimize the policy to maximize the expected return, and a clipping operation is applied to stabilize training: 
\begin{equation}
\begin{aligned}
\mathcal{J}_{\text{PPO}}(\theta) &= \mathbb{E}_{q \sim \mathcal{D}, o_i \sim \pi_{\theta_{\text{old}}}(\cdot|q)} \\
& \left[
     \frac{1}{|o_i|}\sum_{t=1}^{|o_i|} \min\left( r_{i,t}(\theta) \hat{A}_{i,t}, \text{clip}\big(r_{i,t}(\theta), 1 - \epsilon, 1 + \epsilon\big) \hat{A}_{i,t}\right) - \beta D_{KL}(\pi_{\theta} || \pi_{\text{ref}})
\right],
\end{aligned}
\label{eq:grpo_objective} 
\end{equation}
where $r_{i,t}(\theta) = \frac{\pi_\theta(o_{i,t} \mid q, o_{i,<t})}{\pi_{\theta_{\text{old}}}(o_{i,t} \mid q, o_{i,<t})}$ is the importance sampling (IS) weight, which is used to correct the distribution during off-policy updates, because we use $\pi_{\theta_{\text{old}}}$ to collect trajectories to update the policy via multiple steps in a minibatch manner. While PPO requires a separate value model to compute the advantage $\hat{A}_{i,t}$, GRPO~\citep{shao2024deepseekmath} eliminates the value model and defines the advantage as the output reward relative to other responses in the group:
\begin{equation}
\hat{A}_{i,t} = \frac{R_i - \text{mean}(\{R_j\}_{j=1}^G)}{\text{std}(\{R_j\}_{j=1}^G)}, 
\end{equation}
where $R_i$ is the reward of the response, and $G$ responses $\{o_i\}^G_{i=1}$ are sampled for each question. The reward is either from rule-based verifiers such as in mathematical problem solving, or from a reward model. 

\noindent\textbf{Issues of Token Clipping.}
In our initial experiments with the hybrid architecture under the zero-RL setting, we observed that the GRPO algorithm adversely affected training performance and failed to effectively promote the emergence of long CoT reasoning behaviors. Through a series of controlled ablation studies, we ultimately identified the undesirable clipping operation in the original PPO/GRPO loss as the primary factor contributing to degraded learning performance.
Specifically, we found that tokens associated with reflective behaviors (e.g., \texttt{However}, \texttt{Recheck}, \texttt{Wait}, \texttt{Aha}), which often serve as ``forks'' in reasoning paths, were typically rare and assigned low probabilities by our base model. During policy updates, these tokens were likely to exhibit high $r_{i,t}$ values. As a result, these tokens were clipped out after the first on-policy update, preventing them from contributing to subsequent off-policy gradient updates. This issue was particularly pronounced in our hybrid-architecture model and further hindered the scalability of reinforcement learning.
These low-probability tokens, however, are often crucial for stabilizing entropy~\citep{cui2025entropymechanismreinforcementlearning} and facilitating scalable RL~\citep{wang20258020rulehighentropyminority}. Although DAPO attempts to mitigate this issue by increasing the upper clipping bound~\citep{yu2025dapoopensourcellmreinforcement}, we found this approach to be less effective in our setup, which involved 16 rounds of off-policy updates per generation batch.

\begin{figure}[!t]
    \centering
    \includegraphics[width=0.5\textwidth]
    {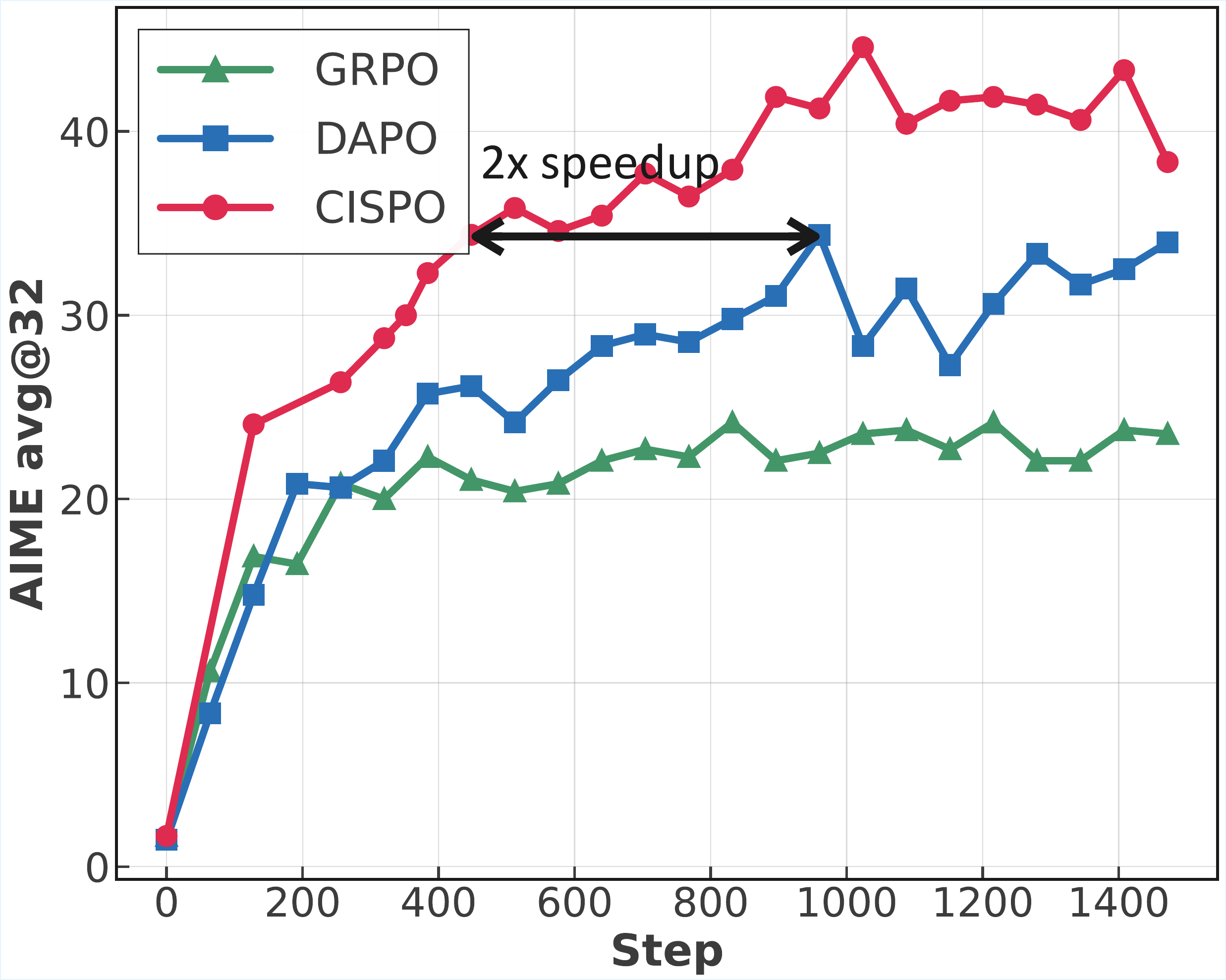}
\caption{Comparison of GRPO, DAPO, and our proposed \method{} on AIME 2024, based on Qwen2.5-32B-base. \method{} outperforms both GRPO and DAPO in terms of performance at the same number of training steps, and achieves comparable performance to DAPO using 50\% of the training steps.}
\label{fig:method-result}
\end{figure}

\noindent\textbf{The \method{} Algorithm.}
In response, we propose a new algorithm that explicitly avoids dropping tokens, even those associated with large updates, while inherently maintaining entropy within a reasonable range to ensure stable exploration. First, recall that the vanilla REINFORCE objective with corrected distribution for offline updates is:

\begin{equation}
    \begin{aligned}
    \mathcal{J}_{\text{REINFORCE}}(\theta) &= \mathbb{E}_{(q,a) \sim \mathcal{D}, o_i \sim \pi_{\theta_{\text{old}}}(\cdot|q)} \\
    & \left[
        \frac{1}{|o_i|} \sum_{t=1}^{|o_i|}
        \texttt{sg}(r_{i,t}(\theta))\hat{A}_{i,t}\log \pi_\theta(o_{i,t} \mid q, o_{i,<t})
    \right], 
    \end{aligned}
\label{eq:reinforce}
\end{equation}
where $\texttt{sg}(\cdot)$ denotes the stop-gradient operation.
Rather than clipping the token updates as in PPO/GRPO, we instead clip the importance sampling weight in Eq.~\ref{eq:reinforce} to stabilize training. 
We term our approach \method{} (\textbf{C}lipped \textbf{IS}-weight \textbf{P}olicy \textbf{O}ptimization). Adopting the group relative advantage from GRPO and the token-level loss~\citep{yu2025dapoopensourcellmreinforcement,liu2025understandingr1zeroliketrainingcritical}, \method{} optimizes the following objective:

\begin{equation}
    \begin{aligned}
    \mathcal{J}_{\text{\method{}}}(\theta) &= \mathbb{E}_{(q,a) \sim \mathcal{D}, \{o_i\}_{i=1}^G \sim \pi_{\theta_{\text{old}}}(\cdot|q)} \\
    & \left[
        \frac{1}{\sum_{i=1}^G |o_i|} \sum_{i=1}^G \sum_{t=1}^{|o_i|}
        \texttt{sg}(\hat{r}_{i,t}(\theta))\hat{A}_{i,t}\log \pi_\theta(o_{i,t} \mid q, o_{i,<t})
    \right], 
    \end{aligned}
\label{eq:CISPO}
\end{equation}
where $\hat{r}_{i,t}(\theta)$ is the clipped IS weight:

\begin{equation}
    \hat{r}_{i,t}(\theta) = \text{clip}\left(r_{
    i,t}(\theta), 1-\epsilon^{IS}_{low}, 1+\epsilon^{IS}_{high}\right).
\end{equation}
We note that without weight clipping, $\mathcal{J}_{\text{\method{}}}$ reduces to the standard policy gradient objective. In our experiments, we did not impose a lower bound on the IS weight by setting $\epsilon^{IS}_{low}$ to a large value; instead, we only tuned $\epsilon^{IS}_{high}$.
Although the gradient of Eq.~\ref{eq:CISPO} is slightly biased due to weight clipping, this approach preserves gradient contributions from all tokens, especially in long responses. 
\method{} proves effective in our experiments, helping reduce variance and stabilizing RL training. 
In addition, we utilize the dynamic sampling and length penalty techniques from~\citet{yu2025dapoopensourcellmreinforcement}. There is no KL penalty term in \method{} similar to other recent works~\citep{yu2025dapoopensourcellmreinforcement,hu2025openreasonerzeroopensourceapproach}.

\noindent\textbf{A General Formulation.}
While we adopt \method{} in our experiments, here we further present a unified formulation by introducing a token-wise mask into the \method{} objective. This allows for hyperparameter tuning to control whether, and under what conditions, gradients from specific tokens should be dropped:

\begin{equation}
    \begin{aligned}
    \mathcal{J}_{\text{unify}}(\theta) &= \mathbb{E}_{(q,a) \sim \mathcal{D}, \{o_i\}_{i=1}^G \sim \pi_{\theta_{\text{old}}}(\cdot|q)} \\
    & \left[
        \frac{1}{\sum_{i=1}^G |o_i|} \sum_{i=1}^G \sum_{t=1}^{|o_i|}
        \texttt{sg}(\hat{r}_{i,t}(\theta))\hat{A}_{i,t}\log \pi_\theta(o_{i,t} \mid q, o_{i,<t})M_{i,t}
    \right]. 
    \end{aligned}
\label{eq:unify}
\end{equation}
The mask $M_{i,t}$ is equivalent to the mask implicitly defined in the PPO trust region:
\begin{equation}
M_{i,t} = 
\begin{cases} 
0 & \text{if } \hat{A}_{i,t} > 0 \text{ and } r_{i,t}(\theta) > 1 + \epsilon_{\text{high}}, \\
0 & \text{if } \hat{A}_{i,t} < 0 \text{ and } r_{i,t}(\theta) < 1 - \epsilon_{\text{low}}, \\
1 & \text{otherwise}.
\end{cases} 
\end{equation}
This unified loss formulation can flexibly represent different clipping strategies under a common framework. 

\noindent\textbf{Empirical Validation of \method{}.}
To validate the effectiveness of \method{}, we empirically compare it with DAPO and GRPO in a zero-RL training setting. Specifically, we apply different RL algorithms to train the Qwen2.5-32B-base model on the mathematical reasoning dataset from~\citet{yu2025dapoopensourcellmreinforcement}, and report performance on the AIME 2024 benchmark. As shown in Figure~\ref{fig:method-result}, \method{} significantly outperforms both DAPO and GRPO with the same number of training steps. Notably, \method{} demonstrates superior training efficiency compared to other approaches; for example, it matches DAPO's performance with only 50\% of the training steps.

\subsection{Efficient RL Scaling with Lightning Attention -- Challenges and Recipes}
\label{sec:challenge}

\begin{figure}[!t]
\begin{subfigure}[t]{0.45\textwidth}  
    \centering
    \includegraphics[width=\textwidth]{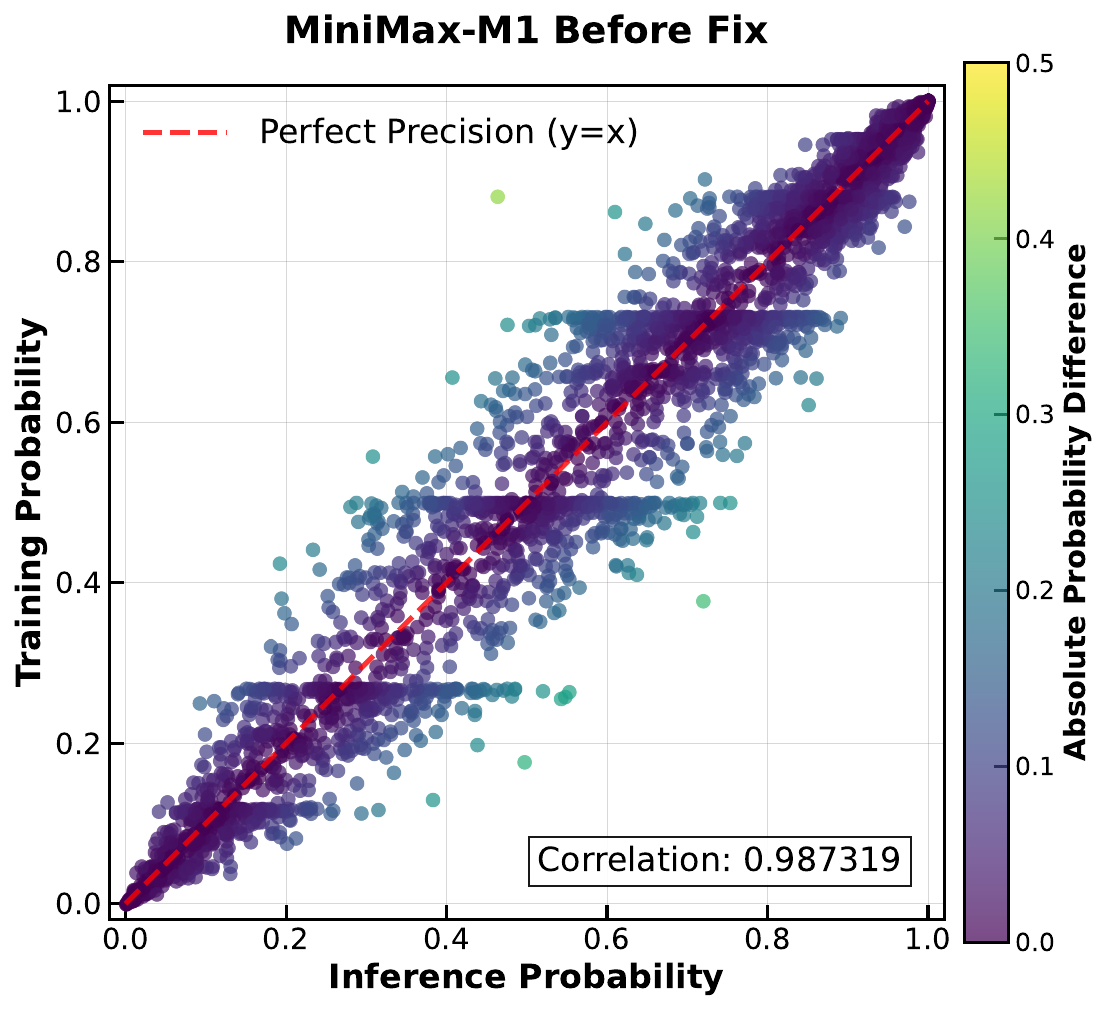}
\end{subfigure}
\hfill
\begin{subfigure}[t]{0.45\textwidth}  
    \centering
    \includegraphics[width=\textwidth]{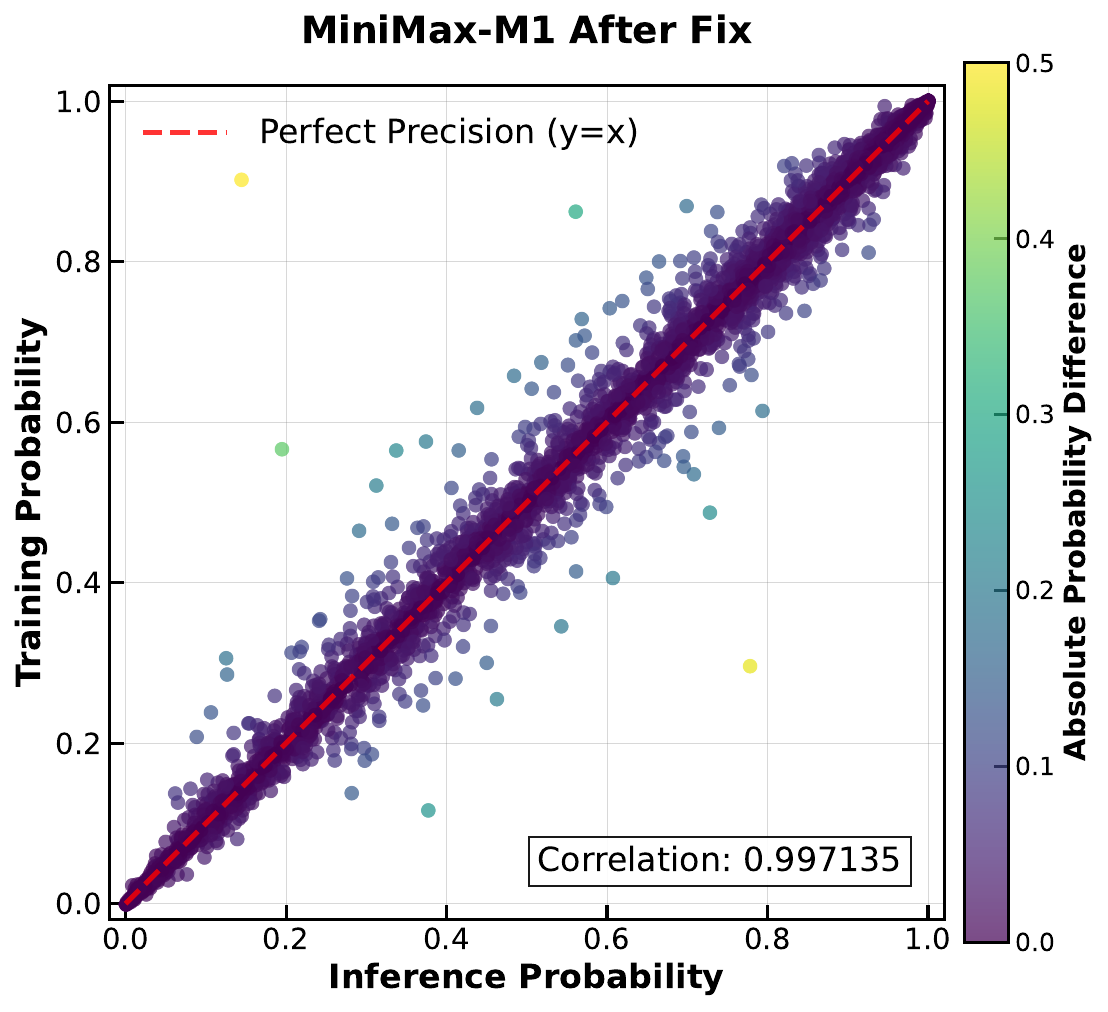}
\end{subfigure}
\caption{Probability of tokens in training-mode code vs. probability of tokens in inference-mode code. Each point in the figures represents an individual token. The Pearson correlation coefficient is indicated in the figures. Theoretically, the two probabilities should be identical, and all the tokens should be exactly on the diagonal line. 
{\bf Left:} Correlation of the M1 model before our fix; {\bf Right:} Correlation of the M1 model after applying our fix of using FP32 precision for the LM output head.}
\label{fig:mis-precision}
\end{figure}

As shown in Figure~\ref{fig:perf-bars-flops} (Right), we emphasize that our hybrid attention inherently enables more efficient RL scaling compared to traditional attention designs, since rollout computation and latency are often the primary bottlenecks in RL training. However, as pioneers in conducting large-scale RL experiments with this novel architecture, we encountered unique challenges and developed targeted solutions, as we describe below.

\noindent \textbf{Computational Precision Mismatch in Generation and Training.} 
RL training is highly sensitive to computational precision. 
During our RL training, we observed a significant discrepancy in the probabilities of rolled-out tokens between training-mode and inference-mode, as shown in Figure~\ref{fig:mis-precision} (Left). This discrepancy arose from a precision mismatch between the training and inference kernels. The issue was detrimental and prevented reward growth in our experiments.
Interestingly, this issue did not appear in smaller, dense models with softmax attention.
Through layer-by-layer analysis, we identified high-magnitude activations in the LM head at the output layer as the primary source of error. To address this, we increased the precision of the LM output head to FP32, thereby realigning the two theoretically identical probabilities, as demonstrated in Figure~\ref{fig:mis-precision} (Right). This adjustment improved the correlation between training and inference probabilities from approximately 0.9x to 0.99x. Notably, this correlation metric remained stable throughout training, enabling successful reward increase.

\noindent \textbf{Optimizer Hyperparameter Sensitivity.}
We employ the AdamW~\citep{loshchilovdecoupled} optimizer, and inappropriate configurations of $\beta_1$, $\beta_2$, and $\epsilon$ can lead to non-convergence during training. ~\cite{molybog2023theoryadaminstabilitylargescale}. For instance, using the default configuration from VeRL~\cite{sheng2024hybridflow}, where betas = (0.9, 0.999) and eps = 1e-8, can result in such issues.
We have observed that the gradient magnitudes in MiniMax-M1 training span a wide range, from 1e-18 to 1e-5, with the majority of the gradients being smaller than 1e-14. Furthermore, the correlation between the gradients of adjacent iterations is weak. Based on this, we set $\beta_1 = 0.9$, $\beta_2 = 0.95$, and eps=1e-15.

\noindent \textbf{Early Truncation via Repetition Detection.}
During RL training, we found that complex prompts could induce pathologically long and repetitive responses, whose large gradients threatened model stability. Our goal was to preemptively terminate these generation loops rather than penalize the already repetitive text. As simple string-matching is ineffective against varied repetition patterns, we developed a heuristic based on token probabilities. We observed that once a model enters a repetitive cycle, the probability for each token soars. Consequently, we implemented an early truncation rule: generation is halted if 3,000 consecutive tokens each have a probability above 0.99. This method successfully prevents model instability and improves generation throughput by eliminating these pathological, long-tail cases.

\section{Scaling Reinforcement Learning with Diverse Data}
\label{sec:data}
In this section, we describe the data and reward we adopted for our RL stage. We incorporate a diverse set of environments in our RL training pipeline, including tasks that can be verified by rules and general tasks that need to be verified through reward models.
All these environments are integrated into the RL stage using a carefully designed curriculum.

\subsection{Reasoning-Intensive Tasks with Rule-based Verification}
Below, we introduce our data that can be verified by deterministic rules. For all the following tasks, we employ rule-based final correctness as the correctness reward, complemented by a format reward.

\noindent \textbf{Mathematical Reasoning.}
Our initial mathematical dataset comprises hundreds of thousands of high-quality, competition-level problems, meticulously curated and organized from public sources and official mathematics competitions. These problems span a wide range of difficulty levels, each paired with a standard reference solution.
Our data cleaning pipeline begins with the removal of incomplete samples and those exhibiting formatting or typographical errors. We subsequently apply embedding-based deduplication across the RL data sources and enforce a strict separation from the SFT dataset to avoid any overlap, as leakage from the SFT phase into the RL stage hinders exploration and undermines training effectiveness. Additionally, we employ both n-gram and embedding-based methods to eliminate potential contamination from commonly used mathematical benchmark test sets, thereby ensuring the integrity and fairness of our evaluations.
We filter out samples containing multiple sub-problems, proof-based questions, and binary questions (e.g., true/false) that are susceptible to random guessing. Multiple-choice questions are reformulated into open-ended formats to better align with our reinforcement learning framework.
Next, we employ our internal model to extract the final answers from the reference solution, retaining only those samples whose extracted answers can be correctly parsed by our rule-based answer checker. Finally, we use a strong reasoning model to compute the pass@10 for each question and retain only those samples with a pass rate strictly between 0 and 0.9, resulting in a curated dataset of nearly 50K high-quality mathematical samples for our RL training.

\noindent \textbf{Logical Reasoning.}
For logical reasoning data, we carefully select 41 logical reasoning tasks requiring non-trivial reasoning ability such as cipher and Sudoku, then we implement a data synthesis framework to synthesize all the data. Concretely, we utilize our SynLogic framework~\citep{liu2025synlogic} to implement the data synthesis pipeline featuring task-specific data generators and rule-based task-specific verifiers, enabling automatic logical data generation. We meticulously configure the difficulty parameters during generation, ensuring the appropriate learning challenge of the generated data. Specifically, to prevent inclusion of overly difficult instances, we establish an upper difficulty bound based on the solvability limits of current strong reasoning models, requiring their pass@10 rates greater than zero. Similarly, we set a lower difficulty bound using the lowest difficulty parameters for which the MiniMax-Text-01 model achieves pass rates between 0 and 0.5. This approach ensures the data maintains a balance between difficulty and learnability. In addition, as the model capabilities improve during training, we increase the difficulty of the data in the later stages. Using this framework, we synthesize approximately 53K logical reasoning samples for RL training.

\noindent \textbf{Competitive Programming.}
For the competitive programming problems, we collect publicly available problems from online judge platforms and popular coding websites. For problems lacking test cases, we develop an LLM-based workflow and use the MiniMax-Text-01 model to generate comprehensive test suites. Similar to our approach with mathematical reasoning datasets, we filter problems based on quality and difficulty using pass rates from model sampling, retaining moderately challenging and high-quality algorithmic problems. Through this process, we generate 30K competitive programming data samples for RL training.

\noindent \textbf{Software Engineering.}
For the software engineering domain, inspired by SWE-bench~\citep{jimenez2024swebenchlanguagemodelsresolve}, we construct verifiable reinforcement learning environments by leveraging real-world data from public GitHub repositories. Our dataset primarily comprises issues and pull requests (PRs) that encapsulate common software development challenges, including bug localization, code repair, and test case synthesis.
To facilitate effective reinforcement learning, we develop a sophisticated containerized sandbox environment that simulates a realistic software development workflow. This environment enables the actual execution of code, providing direct and verifiable feedback on the correctness and efficacy of an agent's proposed interventions. The pass/fail status of pre-defined or newly generated test cases serves as the primary reward signal for our RL framework. A successful execution that passes all relevant test cases yields a positive reward, while compilation errors, runtime failures, or test case regressions result in a zero or negative reward, thus providing a clear signal for policy optimization.
Through this process, we curate several thousand high-quality data samples. Each sample includes a problem description (e.g., bug report from an issue), the initial faulty code, and a set of associated test cases. This setup allows our RL agent to learn to accurately pinpoint bugs, propose correct code fixes, and even synthesize new, effective test cases, with performance directly verifiable through the execution within our sandboxed environment.

\subsection{General Domain Tasks with Model-based Feedbacks}

In this section, we further extend the RL scope to a wider array of general domain tasks. As these tasks cannot be easily verified by rules, we utilize reward models to provide the feedback.

\subsubsection{Data and Reward Models}
Our general RL dataset consists of a total of 25K complex samples. These can be broadly categorized into two types: samples with ground-truth answers that are verifiable but difficult to validate using rules, and samples without ground-truth answers. 

\noindent\textbf{Tasks with Ground Truth.} This category primarily includes STEM and other factual problems where answers are objective but may have multiple valid expressions. Such diversity often renders rule-based answer checkers inaccurate. Our data cleaning process is similar to that used in mathematical reasoning, while we use our Generative Reward Model (GenRM) as a verifier, instead of relying on rule-based checkers.
To evaluate consistency between ground-truth answers and model responses, we adopt a five-grade reward scale to evaluate the two components. First, we construct a human-annotated reward model benchmark, which covers a range of objective tasks across diverse knowledge and task domains, especially the pairs of model response–ground truth that rule-based checkers fail to judge accurately. Second, we evaluate the GenRM's effectiveness by comparing the Best-of-N (BoN) responses selected by GenRM against the pass@N metrics across several benchmarks. GenRM performance is assessed using its accuracy on the human-annotated benchmark and the performance gap between BoN and pass@N. These metrics guide experiments to optimize both the data distribution and the prompt design used during the GenRM training.

\noindent\textbf{Tasks without Ground Truth.} This category encompasses a wider range of tasks, including instruction-following, creative writing, etc. 
Prompts are sampled from a large pool based on our internal tagging system, ensuring a balanced training distribution across fine-grained domains. 
Even though these queries are typically open-ended and do not have a ground-truth answer, we seek to pair a reference answer for each query, which serves as a reference for reward model judgment. To this end, we first generate responses by various internal and external models, and then these reference answers will undergo our internal quality evaluation.
During RL training, we adopt a pairwise comparison framework to evaluate model responses. Each comparison yields a score of -1, 0, or 1, indicating whether the model's output is worse than, similar to, or better than a reference answer. For instruction-following tasks with constraints particularly, we utilize both the rule-based reward to assess whether the response satisfies the constraint, and model-based reward to evaluate response's quality. As with the ground-truth setting, we first build a human-annotated benchmark, incorporating multiple blind preference judgments from reliable annotators. We then refine our scoring criteria and preference prompt to optimize accuracy as well as potential biases, which would be mentioned in \S\ref{sec:genrm-bias} below. To minimize the potential biases, training data are also optimized by several methods, such as multiple-blind consistent judgment, position-switched consistent judgment, etc. Once an optimal GenRM is trained, a Swiss Round scoring system is performed across the training dataset to determine the most suitable reference answer for RL training.

\subsubsection{Addressing Bias of Generative Reward Models for Long CoT}
\label{sec:genrm-bias}

Effective general RL for complex CoT reasoning tasks is critically dependent on accurate and unbiased reward models. Assessing such CoT responses turns out to be challenging, and we found that GenRMs preferred longer outputs over potentially superior concise alternatives, irrespective of actual reasoning quality. This {\bf length bias} is a significant issue as it may substantially misguide RL policy optimization, incentivizing verbosity without substance and inducing reward hacking.
Our initial efforts to improve GenRM fidelity include standard offline strategies: (1) Diversifying training data with a wide range of response lengths, sources, and quality tiers; (2) Incorporating adversarial examples to expose vulnerabilities; and (3) Refining model architectures. However, empirical analysis revealed that purely offline evaluation and preemptive mitigation of length bias in GenRMs frequently failed to prevent length bias during RL training.

Consequently, our core strategy incorporates continuous online monitoring of length bias during RL training. Specific metrics are established to detect whether the RL policy disproportionately extends output lengths to maximize GenRMs rewards without gains in task success or reasoning depth. Upon detecting such detrimental length-seeking behavior, indicative of exploiting GenRMs length bias, immediate GenRMs recalibration is triggered. This iterative adjustment is vital to preempt reward hacking related to output length, ensuring the policy prioritized substantive capability enhancement over superficial text inflation.
Complementing this adaptive approach, RL-side techniques including reward shaping, value clipping, and normalization are systematically employed.
These mechanisms desensitize reward signals to extreme values from superficial characteristics (e.g., length), thereby directing policy optimization toward substantive quality and correctness of its long CoT reasoning.

\subsection{Curriculum of Incorporating Diverse Data}

Given that our RL data spans a wide spectrum of categories, a core challenge is training a single policy capable of excelling on both reasoning-intensive tasks and general domain tasks. 
To address this, our approach entails a carefully managed curriculum and dynamic weighting strategy for reasoning and general-domain tasks during the RL training process with \method{}: we start with only the reasoning-intensive tasks with rule-based reward, and then gradually mix in the general domain tasks. This ensures that the model continues to refine its verifiable skills (e.g., in math and code) while progressively enhancing its performance on a diverse spectrum of general tasks, from complex instruction following to open-ended CoT reasoning.
This mixed RL training encourages the model to learn context-dependent application of its reasoning abilities—applying rigorous, step-by-step deduction for verifiable problems and more flexible, adaptive generation for general queries—all within a unified policy framework. It prevents catastrophic forgetting of specialized skills while fostering broader generalization.

\section{Extending RL Scaling to Longer Thinking}
\label{sec:long-context}
Our first RL training is performed with an output length limit of 40K tokens. Given that the hybrid architecture of M1 natively supports near-linear scaling for longer sequences, as demonstrated in Figure~\ref{fig:perf-bars-flops} (Right), we further extend the generation length during RL training to 80K tokens. This results in a new model, which we refer to as MiniMax-M1-80k.

\noindent\textbf{Data.} 
To efficiently train our RL model for an 80K output length, we utilize our previously trained 40K model to guide the data filtering process. First, we evaluate the pass rates on the curated dataset described in \S\ref{sec:data} and remove samples that are easily solved. We then adjust the data distribution to favor more challenging examples, such as difficult mathematical and coding problems. Additionally, we downsample synthetic reasoning data after observing that it destabilizes long-context RL training. Specifically, outputs generated from this data type often become repetitive and homogenous, and continued exposure to these patterns proves detrimental to the model's overall performance.

\noindent\textbf{Length Scaling Strategy.} To gradually increase the output length, we employ a staged window expansion RL strategy. We begin with an output length of 40K and incrementally expand it to 48K, 56K, 64K, 72K, and ultimately 80K. This staged approach ensures training stability at each step. The transition to a subsequent length is determined by a set of empirical indicators. These include the convergence of perplexity on the generated sequences and whether the 99th percentile of the output lengths is approaching the current context window limit. These signals offer valuable insights into the model's readiness for scaling, which allows us to maintain robust training throughout the process.

\noindent\textbf{Addressing Training Instability During Scaling.} During the scaling process, we encountered a critical issue in the later stages of training at each length window. Specifically, the model exhibited susceptibility to pattern collapse, where the latter portions of generated sequences degraded into incoherent or garbled text. This phenomenon consistently coincided with increased perplexity, indicating compromised generation quality and stability. We identify the root cause: during output length extension, negative samples increase in length substantially faster than positive samples, frequently reaching the context window limit earlier. Consequently, disproportionately large negative gradients accumulate in the latter segments of generation sequences. This imbalance originates from the inherently unequal nature of GRPO's advantage normalization and the token-level loss we adopt.
To address this, we implement three key solutions: (1) Detecting repetitive patterns (consecutive high-probability tokens) with early stopping to prevent excessive context window consumption by repetitive responses; (2) Adopting combined sample-level loss and token-level normalization to alleviate negative-positive sample imbalance and mitigate adverse effects; (3) Decreasing both the gradient clipping threshold and $\epsilon^{IS}_{high}$ to further stabilize generation.

\section{Evaluations} 
\begin{table*}[!t]
\footnotesize
\renewcommand{\arraystretch}{1.5}
\centering
\caption{\textbf{Performance of MiniMax-M1 on core benchmarks.}}
\setlength{\tabcolsep}{0.8mm}
\resizebox{\linewidth}{!}{
\begin{tabular}{c cccc | ccc | cc}
\toprule
\multirow{2}{*}{\textbf{Tasks}}
& \multicolumn{4}{c}{\textbf{Leading Close-Weights Models}}
& \multicolumn{3}{c}{\textbf{Open-Weights Models}}
& \multicolumn{2}{c}{\textbf{Our Models}} \\
\cmidrule(lr){2-5} \cmidrule(lr){6-8} \cmidrule(lr){9-10}
& \makecell{\textbf{OpenAI-o3}}
& \makecell{\textbf{Gemini 2.5} \\\textbf{Pro (06-05)}} 
& \makecell{\textbf{Claude} \\\textbf{4 Opus}}
& \makecell{\textbf{Seed-}\\\textbf{Thinking-}\\\textbf{v1.5}}
& \makecell{\textbf{DeepSeek-}\\\textbf{R1}}
& \makecell{\textbf{DeepSeek-}\\\textbf{R1-0528}}
& \makecell{\textbf{Qwen3-}\\\textbf{235B-A22B}}
& \makecell{\textbf{MiniMax-}\\\textbf{M1-40k}}
& \makecell{\textbf{MiniMax-}\\\textbf{M1-80k}} \\
\makecell{\scriptsize Extended\\\scriptsize Thinking}
& \makecell{\scriptsize \emph{100K}}
& \makecell{\scriptsize \emph{64K}}
& \makecell{\scriptsize \emph{64K}}
& \makecell{\scriptsize \emph{32K}}
& \makecell{\scriptsize \emph{32K}}
& \makecell{\scriptsize \emph{64K}}
& \makecell{\scriptsize \emph{32K}}
& \makecell{\scriptsize \emph{40K}}
& \makecell{\scriptsize \emph{80K}} \\
\midrule
\multicolumn{10}{c}{\emph{Mathematics}} \\
\hline
\makecell{AIME 2024}
& 91.6 & 92.0 & 76.0 & 86.7 & 79.8 & 91.4 & 85.7 & 83.3 & 86.0\\
\makecell{AIME 2025}
& 88.9 & 88.0 & 75.5 & 74.0 & 70.0 & 87.5 & 81.5 & 74.6 & 76.9\\
\makecell{MATH-500}
& 98.1 & 98.8 & 98.2 & 96.7 & 97.3 & 98.0 & 96.2 & 96.0 & 96.8 \\
\hline
\multicolumn{10}{c}{\emph{General Coding}} \\
\hline
\makecell{LiveCodeBench \\\emph{\tiny (24/8$\sim$25/5)}} & 75.8 & 77.1 & 56.6 & 67.5 & 55.9 & 73.1 & 65.9 & 62.3 & 65.0\\
\makecell{FullStackBench}
& 69.3 &   --   & 70.3 & 69.9 & 70.1 & 69.4 & 62.9 & 67.6 & 68.3\\
\hline
\multicolumn{10}{c}{\emph{Reasoning \& Knowledge}} \\
\hline
\makecell{GPQA Diamond}
& 83.3 & 86.4 & 79.6 & 77.3 & 71.5 & 81.0 & 71.1 & 69.2 & 70.0\\
\makecell{HLE \tiny\emph{ (no tools)}}
& 20.3 & 21.6 & 10.7 & 8.2  & 8.6$^*$  & 17.7$^*$ & 7.6$^*$ & 7.2$^*$ & 8.4$^*$\\
\makecell{ZebraLogic}
& 95.8 &  91.6  & 95.1 & 84.4 & 78.7 & 95.1 & 80.3 & 80.1 & 86.8\\
\makecell{MMLU-Pro}
& 85.0 & 86.0 & 85.0 & 87.0 & 84.0 & 85.0 & 83.0 & 80.6 & 81.1\\
\hline
\multicolumn{10}{c}{\emph{Software Engineering}} \\
\hline
\makecell{SWE-bench Verified} & 69.1 & 67.2 & 72.5 & 47.0 & 49.2 & 57.6 & 34.4 & 55.6 & 56.0\\
\hline
\multicolumn{10}{c}{\emph{Long Context}} \\
\hline
\makecell{OpenAI-MRCR} \emph{(128k)}
& 56.5 & 76.8 & 48.9 & 54.3 & 35.8 & 51.5 & 27.7 & 76.1 & 73.4 \\

\makecell{OpenAI-MRCR} \emph{(1M)} & -- & 58.8 & -- & -- & -- & -- & -- & 58.6 & 56.2 \\

\makecell{LongBench-v2} & 58.8 & 65.0 & 55.6 & 52.5 & 58.3 & 52.1 & 50.1 & 61.0 & 61.5 \\
\hline
\multicolumn{10}{c}{\emph{Agentic Tool Use}} \\
\hline
\makecell{TAU-bench \emph{(airline)}} & 52.0 & 50.0 & 59.6 & 44.0  & -- & 53.5 & 34.7 & 60.0 & 62.0\\
\makecell{TAU-bench \emph{(retail)}} & 73.9 & 67.0 & 81.4 & 55.7 & -- & 63.9 & 58.6 & 67.8 & 63.5\\
\hline
\multicolumn{10}{c}{\emph{Factuality}} \\
\hline
\makecell{SimpleQA} & 49.4 & 54.0 & -- & 12.9 & 30.1 & 27.8 & 11.0 & 17.9 & 18.5 \\
\hline
\multicolumn{10}{c}{\emph{General Assistant}} \\
\hline
\makecell{MultiChallenge} & 56.5 & 51.8 & 45.8 & 43.0 & 40.7 & 45.0 & 40.0 & 44.7 & 44.7 \\
\bottomrule
\multicolumn{10}{l}{\footnotesize * conducted on the text-only HLE subset.}
\end{tabular}
}
\label{tab:text-open-source-results-updated} 
\end{table*}

\subsection{Core Benchmarks}

We conduct a comprehensive evaluation of MiniMax-M1 across several key domains: mathematics, general coding, software engineering, reasoning \& knowledge, long context, agentic tool use, factuality, and general assistant ability. We evaluate all tasks using temperature 1.0 and top-p 0.95 sampling.

\begin{itemize}
    \item \textbf{Mathematics:} To evaluate mathematical reasoning capabilities, we utilize several competition level math benchmarks, including MATH-500~\citep{hendrycks2021measuring},  AIME 2024, AIME 2025. For AIME evaluation, we sample 32 times and compute the average passrate as the final score.

     \item \textbf{General Coding:} We assess general programming proficiency using LiveCodeBench~\citep{jainlivecodebench} and FullStackBench~\citep{liu2024fullstackbenchevaluatingllms}, which evaluate code generation across diverse programming tasks. For both benchmarks, we report scores as the average passrate of 16 samples.
     
     \item \textbf{Reasoning \& Knowledge:} We assess domain knowledge and reasoning capabilities through GPQA-Diamond~\citep{rein2024gpqa}, MMLU-Pro~\citep{wang2024mmlu}, and the challenging HLE benchmark~\citep{phan2025humanity}. For GPQA-Diamond, we sample 32 times and report the average passrate.
     For HLE evaluation, we assess the model without external tools. Additionally, we measure logical reasoning ability using ZebraLogic~\citep{lin2025zebralogic}.

    \item \textbf{Software Engineering:} We evaluate software engineering capabilities using SWE-bench Verified~\citep{jimenez2024swebenchlanguagemodelsresolve}, which measures the ability to resolve real-world GitHub issues. We report results derived from the Agentless scaffold~\citep{xia2024agentless}. Departing from the original pipeline, our methodology employs a two-stage localization process (without any embedding-based retrieval mechanisms): initial coarse-grained file localization followed by fine-grained localization to specific files and code elements.
        
    \item \textbf{Long Context:} We evaluate long context understanding using OpenAI-MRCR~\citep{openai_mrcr}, which tests retrieval and disambiguation of multiple similar items within extended contexts, and LongBench-v2~\citep{bai2024longbench2}, a challenging benchmark with 503 multiple-choice questions across contexts ranging from 8k to 2M words.
    
    \item \textbf{Agentic Tool Use:} We assess tool use capabilities through TAU-bench~\citep{yaotau}, which emulates dynamic conversations where agents must utilize API tools while adhering to domain-specific policy guidelines. We evaluate TAU-bench with GPT-4.1 as user model, a general system prompt\footnote{"In each round, you need to carefully examine the tools provided to you to determine if any can be used. You must adhere to all of the policies. Pay attention to the details in the terms. Solutions for most situations can be found within these policies."} and without any custom tools. 
    The maximum number of interaction steps is 40. 

    \item \textbf{Factuality:} To measure factuality of LLMs, we utilize SimpleQA~\citep{wei2024measuring}, an adversarially-collected benchmark of fact-seeking questions with single, indisputable answers.

    \item \textbf{General Assistant:} We evaluate general assistant capabilities using MultiChallenge~\citep{sirdeshmukh2025multichallenge}, which assesses LLMs on conducting realistic multi-turn conversations with human users. We report our scores judged by GPT-4o. 
\end{itemize}

\noindent\textbf{Results on Math, Coding, and other General Tasks.}
 Table~\ref{tab:text-open-source-results-updated} presents our model's performance compared to state-of-the-art large reasoning models. In mathematical reasoning, the MiniMax-M1 models demonstrate strong performance across multiple benchmarks, achieving results comparable to the close-weight model Seed-Thinking-v1.5~\citep{seed2025seed1}. Notably, MiniMax-M1-80k achieves 86.0\% on AIME 2024, placing it second among open-weight models and trailing only the latest DeepSeek-R1-0528 model. For general coding, MiniMax-M1-80k matches Qwen3-235B on LiveCodeBench while outperforming it on FullStackBench, demonstrating robust capabilities among leading open-weight models.
On reasoning \& knowledge benchmarks, MiniMax-M1-80k similarly trails DeepSeek-R1-0528 but achieves competitive performance against other top open-weight models.
On the factuality benchmark SimpleQA, Minimax-M1 models underperform DeepSeek-R1 while outperforming all other open-weight models and Seed-Thinking-v1.5.
On MultiChallenge, both MiniMax models perform comparably to DeepSeek-R1-0528 and Claude 4 Optus, with inferior results only to o3 and Gemini-2.5-Pro.

\noindent\textbf{Highlights in Complex Scenarios: Software Engineering,  Long Context, and Tool use.}
Benefiting from our execution-based, software engineering environments during RL, MiniMax-M1-40k and MiniMax-M1-80k achieve strong scores of 55.6\% and 56.0\% on SWE-bench verified respectively. These results are slightly inferior to DeepSeek-R1-0528's 57.6\% and significantly surpass other open-weights models.
Leveraging its 1M context window, the M1 models significantly outperform all other open-weight models in long-context understanding. They even surpass OpenAI o3 and Claude 4 Opus, ranking second globally and trailing only Gemini 2.5 Pro by a small margin.
In agentic tool-use scenarios (TAU-bench), MiniMax-M1-40k surpasses all open-weight models and even Gemini-2.5 Pro. 
Moreover, MiniMax-M1-80k consistently outperforms MiniMax-M1-40k across most benchmarks, confirming the benefits of scaling test-time compute.

\begin{figure*}[t]
\begin{subfigure}[t]{0.32\textwidth}  
    \centering
    \includegraphics[width=\textwidth]{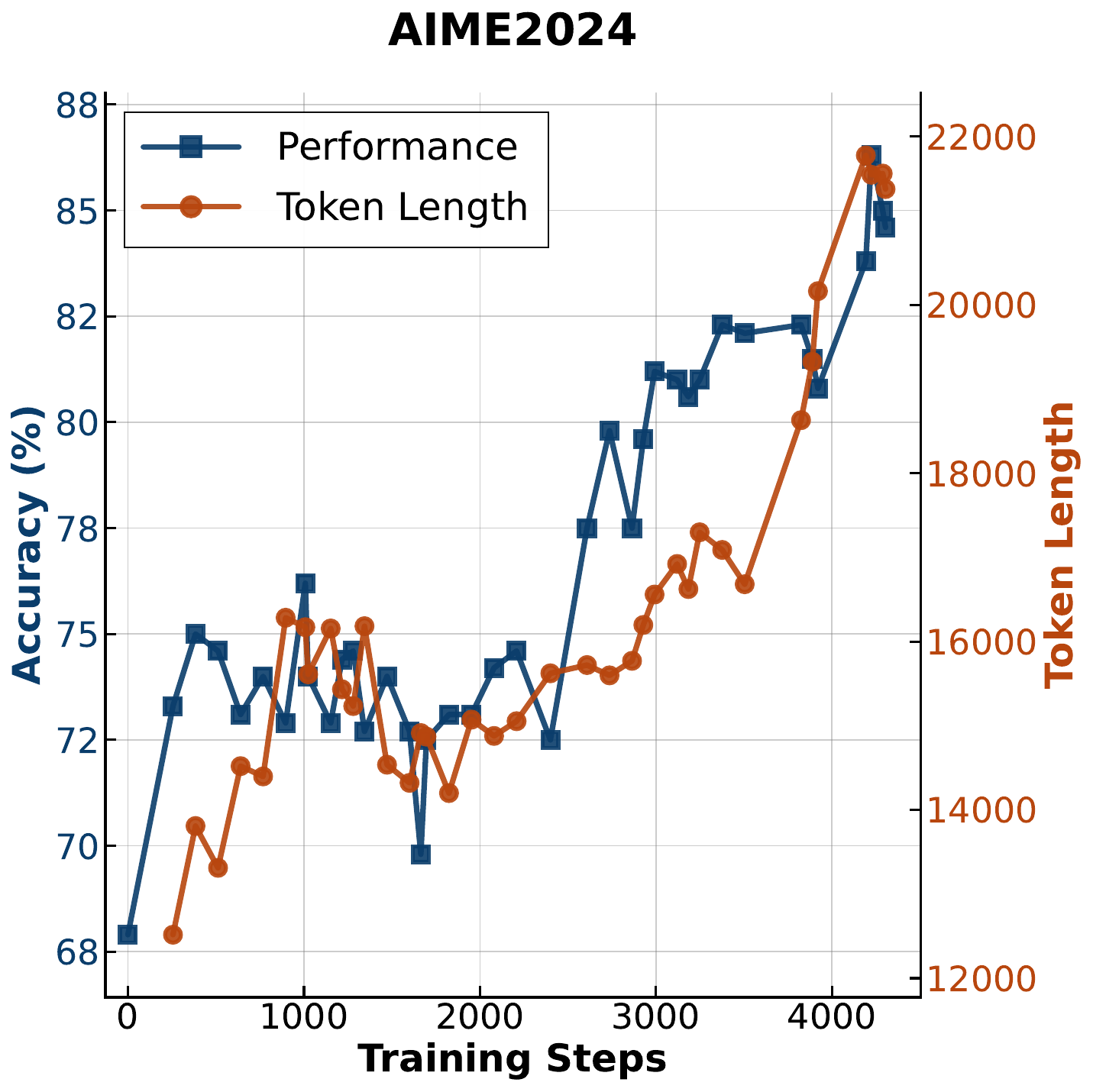}
\end{subfigure}
\hfill
\begin{subfigure}[t]{0.32\textwidth}  
    \centering
    \includegraphics[width=\textwidth]{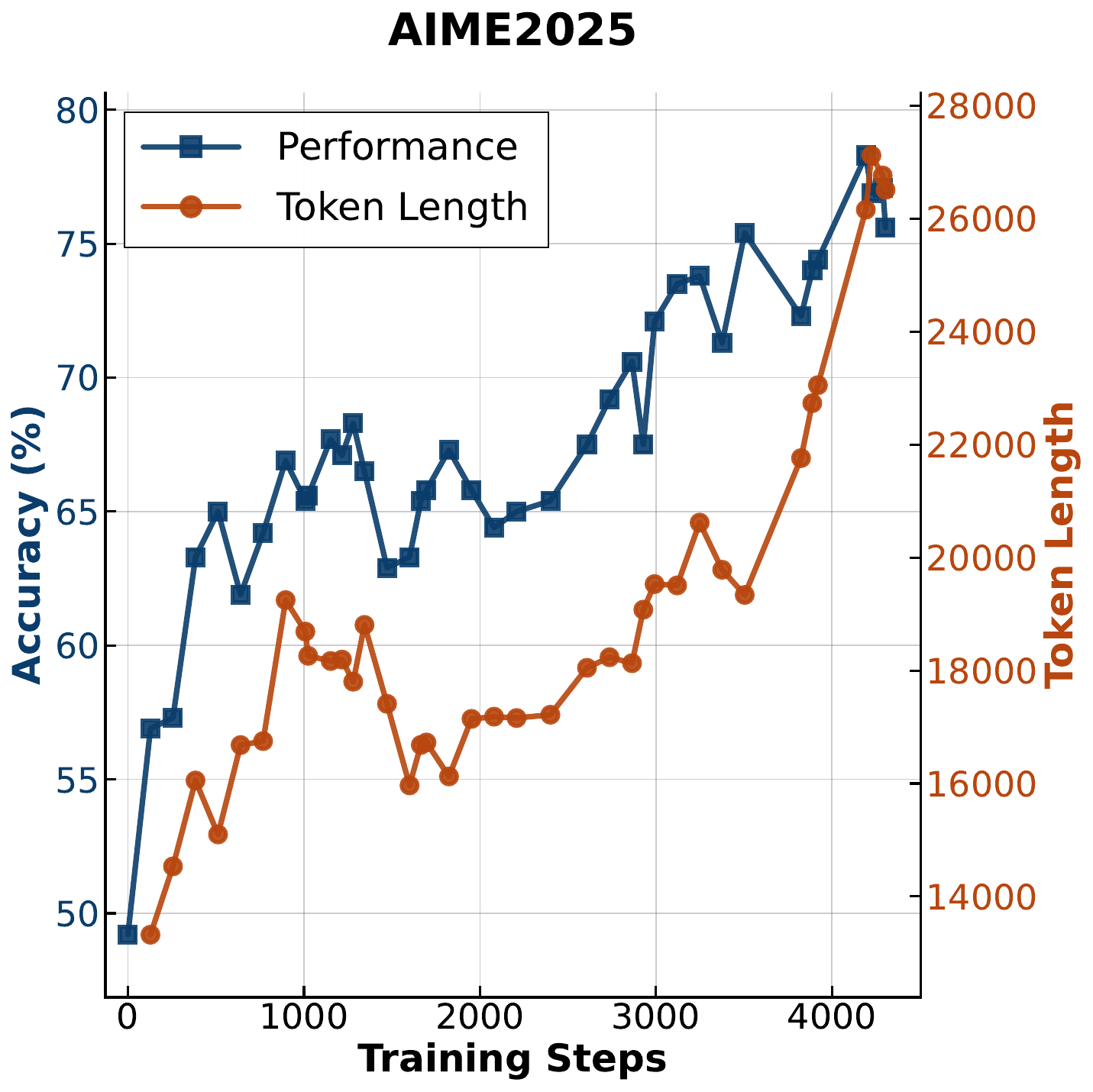}
\end{subfigure}
\hfill
\begin{subfigure}[t]{0.32\textwidth}  
    \centering
    \includegraphics[width=\textwidth]{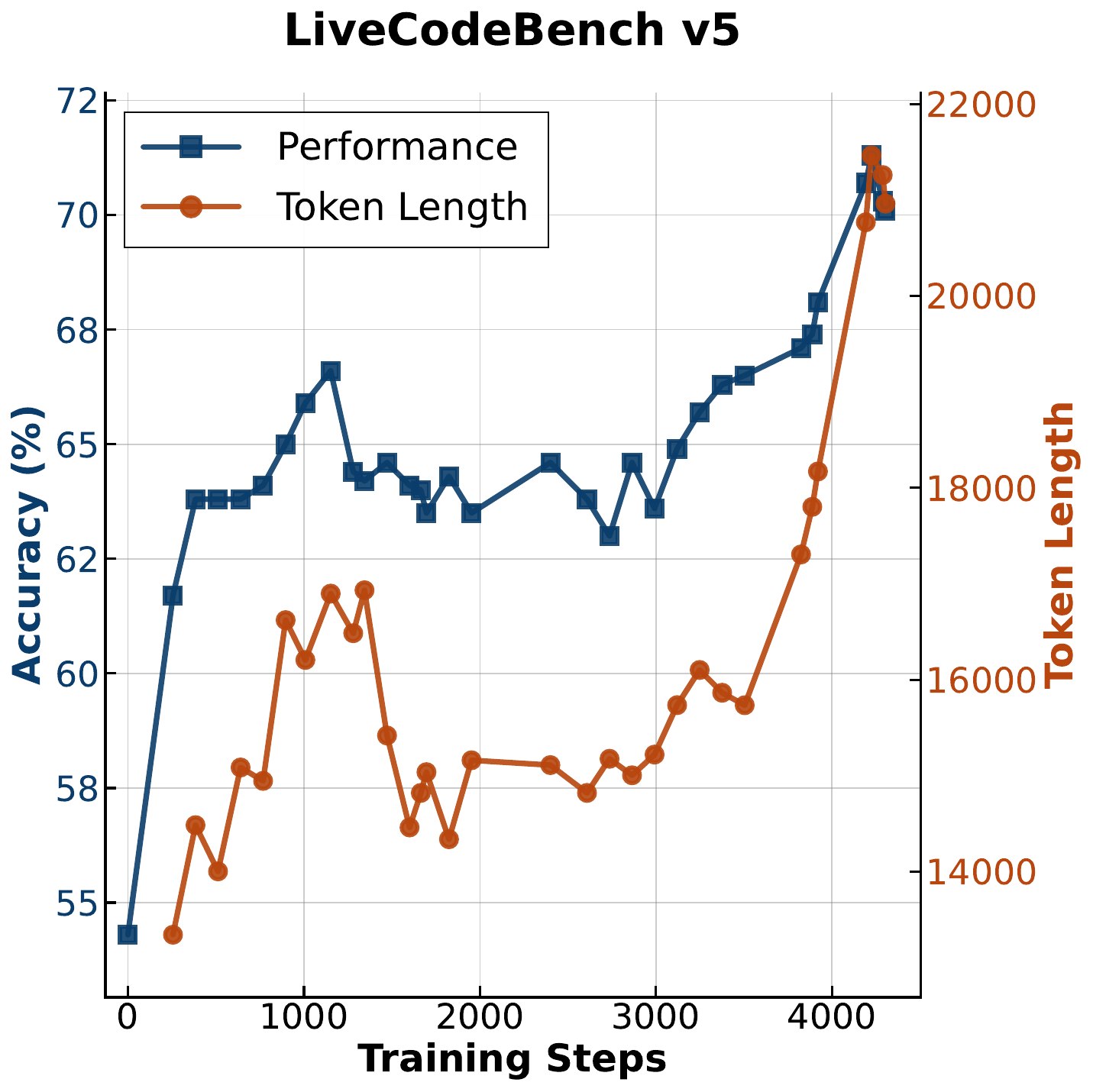}
\end{subfigure}
\caption{Accuracy and generation length versus RL training steps for MiniMax-M1.}
\label{fig:scaling_results}
\end{figure*}

\subsection{Effect of RL Scaling}

To investigate the effect of RL scaling, we track performance and response length throughout training. Figure~\ref{fig:scaling_results} presents three representative examples from AIME 2024, AIME 2025, and LiveCodeBench v5, respectively. 
We observe consistent improvements in both model performance and response length during training. Notably, average response lengths on AIME and LiveCodeBench exceed 20,000 tokens, with AIME 2024 accuracy showing substantial gains from 68\% to 80\%. Crucially, the strong correlation between accuracy gains and increased response length in these visualizations underscores the importance of extending RL scaling to facilitate more extensive reasoning processes.

\section{Conclusion and Future work}
In this work, we introduce and release MiniMax-M1, the world's first open-weight, large-scale reasoning model featuring a lightning attention mechanism. This efficient attention design enables MiniMax-M1 to natively support inputs of up to 1M tokens and generation lengths of 80K tokens---both significantly exceeding capabilities of other open-weight models. These capabilities render MiniMax-M1 uniquely suited for complex, realistic scenarios requiring long context and extended reasoning, properties empirically validated by its strong performance on software engineering, agentic tool use, and long-context understanding benchmarks.
Beyond the inherent efficiency advantages of lightning attention for RL training, this work contributes a novel RL algorithm, \method{}, to accelerate training. Combining architectural advantages with \method{}, we efficiently trained MiniMax-M1, with complete RL training completed in three weeks using 512 H800 GPUs. Across comprehensive evaluations, MiniMax-M1 ranks among the world's best open-weight models alongside DeepSeek-R1 and Qwen3-235B.

Looking forward, as test-time compute continuously scales to power increasingly complex scenarios, we foresee significant potential for such efficient architectures in addressing real-world challenges. These include automating company workflows~\citep{xu2025theagentcompanybenchmarkingllmagents} and conducting scientific research~\citep{si2024llmsgeneratenovelresearch,openai2025research}. Real-world applications particularly demand LRMs that function as agents interacting with environments, tools, computers, or other agents---requiring reasoning across dozens to hundreds of turns while integrating long-context information from diverse sources. We envision MiniMax-M1 serving as a strong foundation for such applications with unique advantages, and we are fully dedicated to further evolving MiniMax-M1 toward this goal.

\bibliography{sample_uniform_arxiv}

\newpage 
\appendix 

\newpage

\section{Contributors} 
The contributors to the report are listed in alphabetical order as follows:

Aili Chen,
Aonian Li,
Bangwei Gong,
Binyang Jiang,
Bo Fei,
Bo Yang,
Boji Shan,
Changqing Yu,
Chao Wang,
Cheng Zhu,
Chengjun Xiao,
Chengyu Du,
Chi Zhang,
Chu Qiao,
Chunhao Zhang,
Chunhui Du,
Congchao Guo,
Da Chen,
Deming Ding,
Dianjun Sun,
Dong Li,
Enwei Jiao,
Haigang Zhou,
Haimo Zhang,
Han Ding,
Haohai Sun,
Haoyu Feng,
Huaiguang Cai,
Haichao Zhu,
Jian Sun,
Jiaqi Zhuang,
Jiaren Cai,
Jiayuan Song,
Jin Zhu,
Jingyang Li,
Jinhao Tian,
Jinli Liu,
Junhao Xu,
Junjie Yan,
Junteng Liu,
Junxian He,
Kaiyi Feng,
Ke Yang,
Kecheng Xiao,
Le Han,
Leyang Wang,
Lianfei Yu,
Liheng Feng,
Lin Li,
Lin Zheng,
Linge Du,
Lingyu Yang,
Lunbin Zeng,
Minghui Yu,
Mingliang Tao,
Mingyuan Chi,
Mozhi Zhang,
Mujie Lin,
Nan Hu,
Nongyu Di,
Peng Gao,
Pengfei Li,
Pengyu Zhao,
Qibing Ren,
Qidi Xu,
Qile Li,
Qin Wang,
Rong Tian,
Ruitao Leng,
Shaoxiang Chen,
Shaoyu Chen,
Shengmin Shi,
Shitong Weng,
Shuchang Guan,
Shuqi Yu,
Sichen Li,
Songquan Zhu,
Tengfei Li,
Tianchi Cai,
Tianrun Liang,
Weiyu Cheng,
Weize Kong,
Wenkai Li,
Xiancai Chen,
Xiangjun Song,
Xiao Luo,
Xiao Su,
Xiaobo Li,
Xiaodong Han,
Xinzhu Hou,
Xuan Lu,
Xun Zou,
Xuyang Shen,
Yan Gong,
Yan Ma,
Yang Wang,
Yiqi Shi,
Yiran Zhong,
Yonghong Duan,
Yongxiang Fu,
Yongyi Hu,
Yu Gao,
Yuanxiang Fan,
Yufeng Yang,
Yuhao Li,
Yulin Hu,
Yunan Huang,
Yunji Li,
Yunzhi Xu,
Yuxin Mao,
Yuxuan Shi,
Yuze Wenren,
Zehan Li,
Zelin Li,
Zhanxu Tian,
Zhengmao Zhu,
Zhenhua Fan,
Zhenzhen Wu,
Zhichao Xu,
Zhihang Yu,
Zhiheng Lyu,
Zhuo Jiang,
Zibo Gao,
Zijia Wu,
Zijian Song,
Zijun Sun

\end{document}